\documentclass[letterpaper, 10 pt, conference]{ieeeconf}  
\IEEEoverridecommandlockouts
\usepackage{graphicx}
\usepackage{amsmath}
\usepackage{amssymb}
\usepackage{xcolor}
\usepackage{booktabs}
\usepackage{subcaption}

\title{\bf Category-Level 6D Object Pose Estimation in Agricultural Settings Using a Lattice-Deformation Framework and Diffusion-Augmented Synthetic Data}

\author{Marios Glytsos$^{1}$, Panagiotis P. Filntisis$^{1}$, George Retsinas$^{1}$, Petros Maragos$^{1,2}$
\thanks{$^{1}$ M. Glytsos, P. P. Filntisis and G. Retsinas are with the Robotics Institute, Athena Research and Innovation Center, Maroussi 15125, Greece}
\thanks{$^{2}$ P. Maragos is with the Robotics Institute, Athena Research and Innovation Center, Maroussi 15125, Greece and the School of ECE, National Technical University of Athens, Greece}
}

\begin{document}

\maketitle
\thispagestyle{empty}
\pagestyle{empty}

\begin{abstract}

Accurate 6D object pose estimation is essential for robotic grasping and manipulation, particularly in agriculture, where fruits and vegetables exhibit high intra-class variability in shape, size, and texture. The vast majority of existing methods rely on instance-specific CAD models or require depth sensors to resolve geometric ambiguities, making them impractical for real-world agricultural applications. In this work, we introduce \textbf{PLANTPose}, a novel framework for category-level 6D pose estimation that operates purely on RGB input. PLANTPose predicts both the 6D pose and deformation parameters relative to a base mesh, allowing a single category-level CAD model to adapt to unseen instances. This enables accurate pose estimation across varying shapes without relying on instance-specific data. To enhance realism and improve generalization, we also leverage Stable Diffusion to refine synthetic training images with realistic texturing, mimicking variations due to ripeness and environmental factors and bridging the domain gap between synthetic data and the real world. 
Our evaluations on a challenging benchmark that includes bananas of various shapes, sizes, and ripeness status demonstrate the effectiveness of our framework in handling large intraclass variations while maintaining accurate 6D pose predictions, significantly outperforming the state-of-the-art RGB-based approach MegaPose. 
\end{abstract}

\section{INTRODUCTION}

Robotics is transforming agriculture by offering scalable solutions for automated harvesting, reducing labor costs, and improving efficiency\cite{kootstra2021selective,zhang2022algorithm}. At the core of these advances lies robotic grasping, which requires accurate 6D pose estimation of fruits and vegetables to enable precise picking and handling. However, unlike rigid industrial objects, fruits and vegetables exhibit significant natural variations in shape, size, and texture not only across different types but also within the same category due to growth, ripeness, and environmental conditions, making pose estimation particularly challenging.

\begin{figure}
    \centering
    \includegraphics[width=1.0\linewidth]{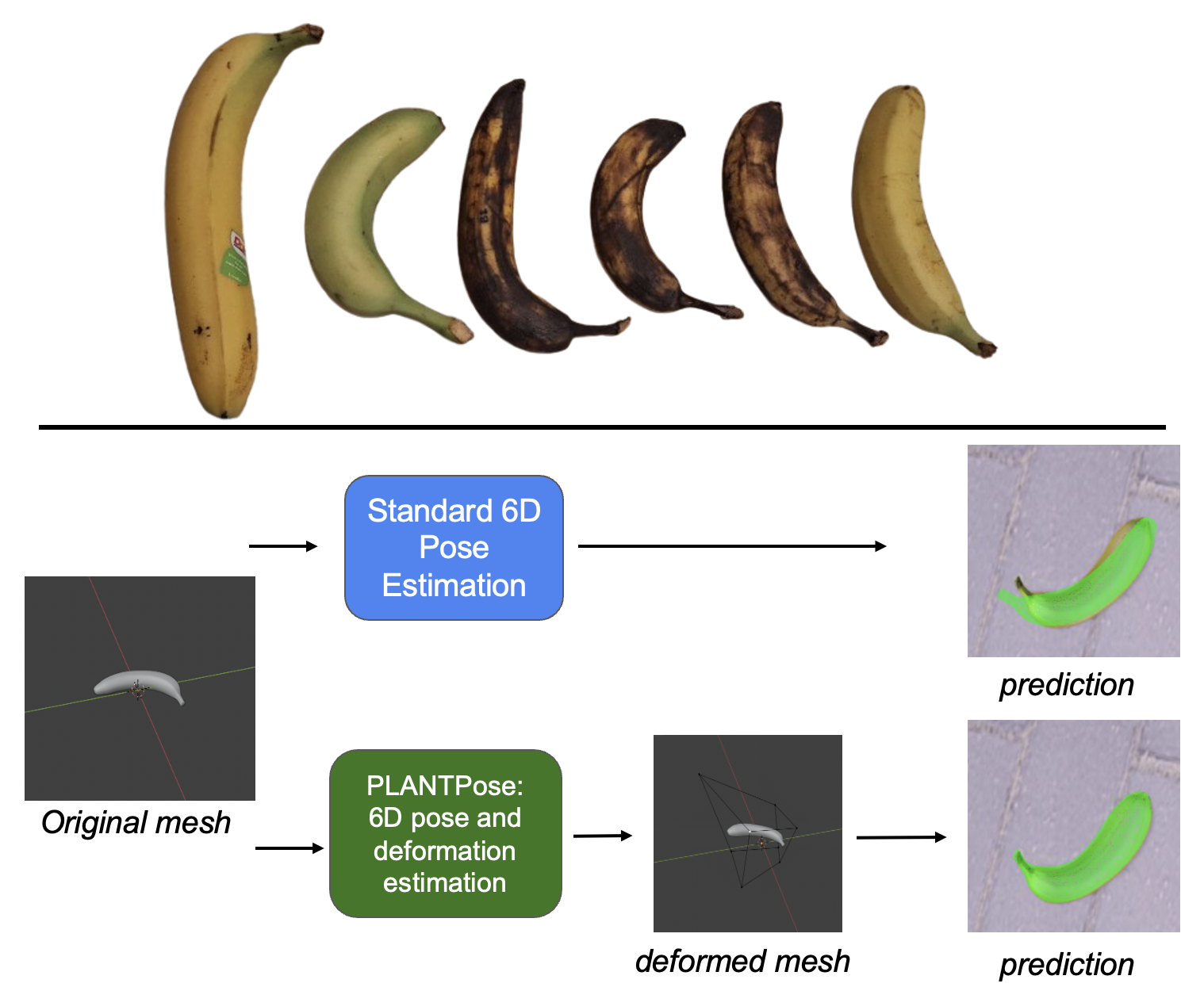}
    \caption{Fruits exhibit significant intra-class variability in shape, size, and texture, posing challenges for standard 6D pose estimation methods. PLANTPose addresses this by simultaneously predicting both the deformation of a base mesh and the 6D pose, enabling more accurate and adaptable predictions.}
    \label{fig:story}
\end{figure}

Most existing 6D pose estimation methods rely on instance-level models, where a specific object must be known beforehand—either as a detailed CAD model \cite{labbe2022megapose,foundation_pose,posecnn,cosypose,nguyen2024gigapose} or through multiple reference images \cite{foundation_pose}. While effective in structured environments, these approaches are impractical for agriculture, where fruits and vegetables naturally vary in form, are encountered in novel configurations, and often appear amidst dense foliage or in close proximity to other produce. A more scalable alternative is category-level pose estimation, which generalizes across object instances within a class without requiring exact geometric models \cite{nocs,socs}. However, current methods often depend on depth sensors to resolve geometric ambiguities, require complex non-differentiable solvers, or involve computationally expensive iterative refinements, limiting their deployment in real-world agricultural applications and purely RGB environments. As a result, achieving robust category-level pose estimation from RGB images alone remains challenging, particularly when large shape variability and real-time performance constraints are involved.

Traditionally, 6D pose estimation methods rely on synthetic data for training, using simulated environments to generate large-scale datasets \cite{bop_datasets,shapenet,google_scanned_objects,objaverse}. While synthetic datasets provide controlled and scalable training, they often fail to capture the full spectrum of real-world variations, a limitation that is particularly pronounced in agricultural products. Fruits not only vary in shape and size but also undergo significant visual changes as they ripen, making robust 6D pose estimation even more challenging. Moreover, collecting a comprehensive set of 3D models and textures for even a single object category is labor-intensive, impractical, and often infeasible.

To address these shortcomings, we introduce \textbf{PLANTPose} (\textbf{P}ose estimation using \textbf{L}attice deform\textbf{A}tio\textbf{N} for ca\textbf{T}egories), a novel framework for 6D pose estimation that employs intuitive deformations on base meshes to adapt a single category-level CAD model to unseen instances (see Fig. \ref{fig:story}). By leveraging a compact set of deformation parameters, our method can capture broad intra-class shape variations \textit{without the need for instance-specific models or depth data}. This makes PLANTPose particularly well-suited for categories such as agricultural produce, where high shape variability is the norm.
\textbf{To further enhance realism and improve generalization to real-world imagery}, we employ Stable Diffusion-based image inpainting on our synthetic datasets, augmenting them with more realistic texturing to better mimic the diverse appearance of fruits at different stages of ripeness. This significantly narrows the domain gap between synthetic and real images, boosting pose estimation accuracy in real-world scenarios. We validate PLANTPose on the banana fruit, which naturally exhibits substantial shape and size deformations and undergoes significant visual changes as it transitions from unripe to ripe and eventually to rotten. This makes bananas an ideal test case for evaluating our framework. Extensive experiments demonstrate that PLANTPose achieves high pose estimation accuracy while generalizing well across diverse shapes, sizes, and textures, highlighting its effectiveness for real-world agricultural applications.

In short, our key contributions are:
\begin{itemize}
    \item We present a novel framework for category-level 6D pose estimation of agricultural produce, leveraging intuitive deformations for flexible shape adaptation.
    \item We enhance the realism of synthetic datasets by leveraging Stable Diffusion with curated prompts and depth conditioning to capture the diverse textures of fruits at different ripeness stages.
    \item We compare our method against a widely used state-of-the-art 6D pose estimation approach, demonstrating that PLANTPose achieves significantly higher accuracy while effectively generalizing to diverse intra-class variations in fruit shapes and textures.
\end{itemize}

The source code and the synthetic dataset will be made publicly available.

\begin{figure*}[t]
    \centering
    \includegraphics[width=1.0\textwidth]{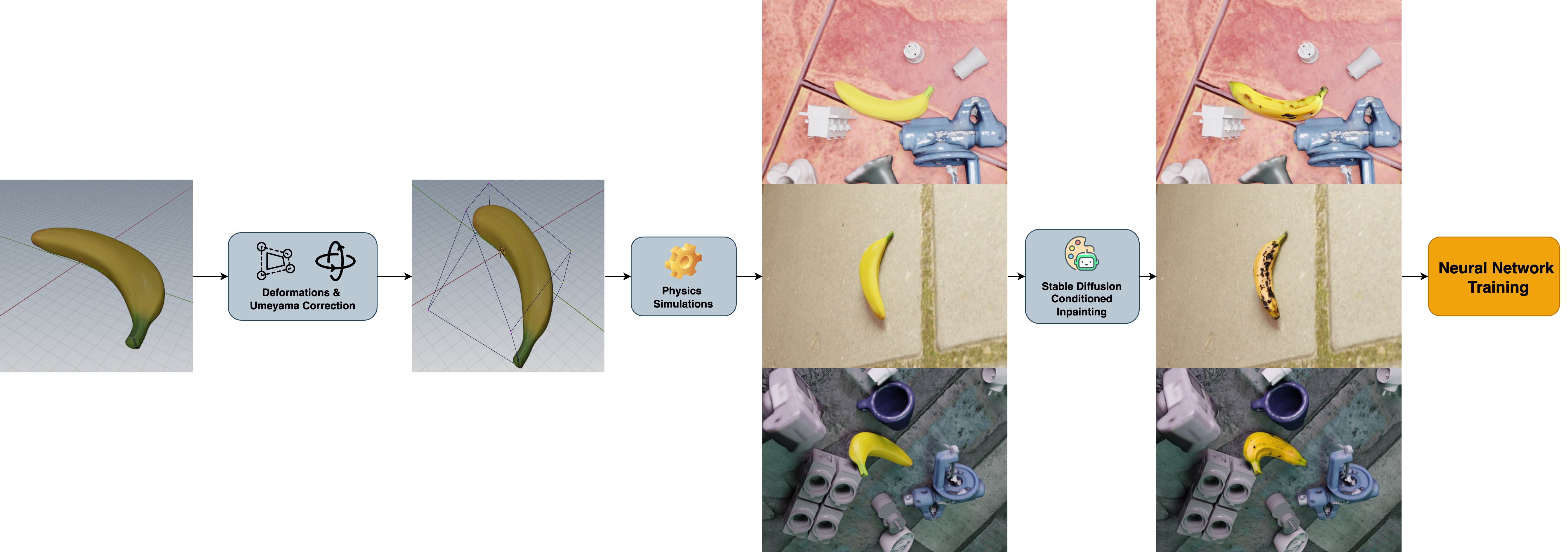}
    \caption{Overview of the PLANTPose framework. Given a base mesh, we generate deformed object instances using lattice-based deformations. These instances are placed in physically plausible synthetic scenes rendered in Blender, which are further refined using Stable Diffusion inpainting to enhance texture realism. The generated dataset is then used to train a deep learning model for simultaneous 6D pose estimation and deformation prediction.}
    \label{fig:pipeline}
\end{figure*}

\section{Related Work}

\textbf{Category-level 6D pose estimation} Category-level approaches predict the pose of previously unseen objects within a defined category. Many works adopt categorical mean shapes to facilitate feature alignment, improving robustness under intra-class variation. One of the early methods \cite{nocs}, introduced a canonical object space where point correspondences between input images and a normalized coordinate system are used to estimate pose. However, it struggled with large shape variations, leading to extensions such as SOCS \cite{socs}, which introduces semantically-aware keypoint alignment, and NuNOCS \cite{nunocs}, which supports non-uniform scaling for objects with varying aspect ratios. Several methods utilize shape priors to handle intra-class variations effectively. Tian et al. \cite{shape_prior_deform} proposed learning a categorical shape prior via an autoencoder and deforming it to match observed instances. Zhang et al. \cite{ssp} improved upon this by introducing symmetry-aware shape prior deformation , allowing for direct pose regression while mitigating ambiguity in symmetric objects.

Implicit representations have also become popular in 6D pose estimation by providing continuous, differentiable shape modeling. ShAPO \cite{shapo} jointly predicts shape, pose, and size using learned implicit fields, while \cite{disp6d} DISP6D separates shape and pose into distinct latent spaces for improved generalization. Neural Radiance Fields (NeRF) \cite{nerf} have also been explored for category-level pose estimation, with approaches like NeRF-Pose \cite{nerfpose} reconstructing object geometry before estimating pose. Unlike traditional correspondence-based methods, these techniques work in a continuous space, reducing alignment errors caused by discrete feature matching.

Recent advancements have focused on generalizing pose estimation beyond specific categories. FoundationPose \cite{foundation_pose} demonstrates strong performance on novel objects. Unlike previous category-specific methods, FoundationPose bridges model-based and model-free approaches by using implicit neural representations for novel view synthesis. While powerful, it still requires CAD models or multiple reference images, limiting its adaptability to truly unconstrained environments. 

\textbf{Synthetic Data for 6D Pose Estimation} Many of the previous works in 6D pose estimation heavily rely on synthetic data due to the difficulty and time-consuming process of annotating real images or video datasets \cite{foundation_pose}. Most methods use CAD models or mesh models available in large scale 3D model databases \cite{shapenet, google_scanned_objects, objaverse}, which store 3D object geometry with vertices, faces, and relative scale, making them suitable for rendering. To generate datasets, these models are placed in synthetic environments, such as Blender \cite{blenderproc}, where images can be rendered with ground truth annotations for position and rotation. This approach allows for large-scale dataset creation without manual annotation. Additionally, Anagnostopoulou et al. \cite{anagnostopoulou2023realistic} leveraged Stable Diffusion with ControlNet to generate highly realistic synthetic data for mushrooms, demonstrating the effectiveness of diffusion-based approaches in modeling agricultural produce. In contrast to previous methods that rely solely on synthetic data, our generation pipeline combines traditional rendering with Stable Diffusion. We first generate physically plausible poses and scenes using physics-based simulation, and then enhance the results via img2img Stable Diffusion inpainting.

\textbf{6D Pose Estimation for Agriculture}
Given the growing importance of 6D pose estimation in agricultural harvesting, numerous methods now address this challenge. Retsinas et al. \cite{retsinas2023mushroom} employed fully convolutional networks with implicit pose encodings to jointly perform mushroom segmentation and pose estimation, while Deep-ToMaTOS \cite{KIM2022107300} introduced a deep learning framework that simultaneously predicts a tomato’s ripeness level and its 6D pose. Li et al. \cite{10802107} further proposed a method for estimating both the 6D pose and 3D scale of strawberries, leveraging synthetic data and a YOLO-based architecture. Costanzo et al. \cite{costanzo2023enhanced} presented a system for apple grasping with an RGB-D camera, and an evaluation of different 6D pose estimation algorithms using RGB-D inputs for grasping was conducted in \cite{robotics13090127}.

Although these approaches specifically target agricultural tasks, they primarily rely on adaptations of standard 6D pose pipelines and do not explicitly model object deformations. In contrast, our method is the first to estimate both the 6D pose and per-instance deformations—a crucial step toward accurately handling the natural shape variability of fruits and vegetables in real-world harvesting scenarios.

\section{Method}
The core concept behind PLANTPose is illustrated in Figure~\ref{fig:pipeline}. Starting with a base mesh, we generate various deformations using lattice-based modeling (Sec.~\ref{sec:lattice}). Next, we create physically plausible synthetic data with Blender, which is further refined using Stable Diffusion to enhance the texture realism of the object of interest (Sec.~\ref{sec:synthetic}). Finally, we train a deep learning model on the resulting dataset (Sec.~\ref{sec:training}). The following sections describe each step in detail.

\subsection{Lattice Deformations}
\label{sec:lattice}
In order to model intuitively the deformations around the object, we use lattices.
In computational geometry, lattices serve as a framework for space partitioning, enabling efficient representation and transformation of spatial data \cite{toth2017handbook}. 

The concept of lattice-based deformations was introduced in computer graphics to provide a structured approach for manipulating 3D objects smoothly. A lattice consists of a set of control points arranged in a grid, where each point influences the space around it. 
The term "lattice" in this context refers to its structured, grid-like nature, which allows for spatial transformations. 
The deformation of any given point inside the lattice is controlled by interpolation between control points.
In our case, the bounding box of the object is the lattice that controls the deformation. 
This way, we control the deformation by the 3D movement of the bounding box boundaries (i.e., the control points). 
In other words, we can simulate useful deformations using only $8 \times 3$ parameters. 
Despite its simplicity, such an approach is able to create non-trivial variations while assisting the formulation of the estimation step, since the developed neural network \textbf{can detect deformation as a regressor of a fixed-sized feature}. To avoid extreme, non-useful cases, we constraint the magnitude of the deformations to an empirical fixed upper bound.

Having defined the control points, one should perform the deformation in the lattice interior using an interpolation step, typically \emph{linear interpolation} or \emph{B-spline interpolation}. We focus on \emph{cubic B-spline} interpolation,  which provides smoother (\(C^2\)) deformations.

Let \(\mathbf{box\_min} = (\mathrm{min}_x, \mathrm{min}_y, \mathrm{min}_z)\) and 
\(\mathbf{box\_max} = (\mathrm{max}_x, \mathrm{max}_y, \mathrm{max}_z)\) be the bounding box corners.
A mesh vertex \(\mathbf{p} = (p_x, p_y, p_z)\) is mapped to \(\mathbf{u} = (u, v, w) \in [0,1]^3\)  so each \((u,v,w)\) is the fractional distance along the \(x,y,z\) axes. A cubic B-spline lattice is determined by eight corner offsets 
\(\mathbf{C}_{i,j,k}\in \mathbb{R}^3\) for \((i,j,k)\in \{0,1\}^3\). 
These corner offsets can be extended (clamped) to a \(4\times 4\times 4\) grid for correct boundary behavior.

The 1D B-spline basis functions for \(t \in [0,1]\) are
\begin{equation}
{\small
   W_0(t) = \frac{1 - 3t + 3t^2 - t^3}{6}, \quad
   W_1(t) = \frac{4 - 6t^2 + 3t^3}{6}
}
\end{equation}
\begin{equation}
W_2(t) = \frac{(1 + 3t + 3t^2 - 3t^3)}{6}, \quad
W_3(t) = \frac{t^3}{6}.
\end{equation}
They are evaluated at \(u,v,w\) along the \(x,y,z\) axes, respectively, giving 
\(W_i(u), W_j(v), W_k(w)\).
A vertex's displacement is computed by summing over the three coordinate axes separately.  
Let each control point \(\mathbf{C}_{i,j,k}\) have components 
\(\bigl(C_{i,j,k}^x,\,C_{i,j,k}^y,\,C_{i,j,k}^z\bigr)\). 
Then, for \((u,v,w)\in[0,1]^3\),
\[
\delta_x(u,v,w)
=
\sum_{i=0}^{3}\sum_{j=0}^{3}\sum_{k=0}^{3}
W_i(u)\,W_j(v)\,W_k(w)\;C_{i,j,k}^x,
\]
\[
\delta_y(u,v,w)
=
\sum_{i=0}^{3}\sum_{j=0}^{3}\sum_{k=0}^{3}
W_i(u)\,W_j(v)\,W_k(w)\;C_{i,j,k}^y,
\]
\[
\delta_z(u,v,w)
=
\sum_{i=0}^{3}\sum_{j=0}^{3}\sum_{k=0}^{3}
W_i(u)\,W_j(v)\,W_k(w)\;C_{i,j,k}^z.
\]
These form the displacement vector 
\(\boldsymbol{\delta}(u,v,w) = \bigl[\delta_x,\delta_y,\delta_z\bigr]\), and the 
deformed position becomes
\(\mathbf{p}' = \mathbf{p} + \boldsymbol{\delta}(u,v,w)\).

Because B-splines ensure \(C^2\) continuity in each dimension, 
they produce more natural, smoothly varying deformations, 
making them preferable for applications requiring realistic 
shape modifications.

\subsection{Synthetic Data Generation}
\label{sec:synthetic}

Creating a real large-scale dataset that adequately captures intra-class variations with deformation data and includes accurate 3D annotations of the pose/deformation is infeasible in practice.
Therefore, we rely on synthetic data generation to cover a wide range of shape and pose variations. 
For this, we use BlenderProc \cite{blenderproc}, a Python-based API for Blender, to simulate physically plausible object placement and rendering.

Our synthetic dataset is generated by constructing virtual environments with 3D object models, where target objects are placed alongside distractors to introduce occlusions. Scenes include procedural room geometry (walls, floors) for context, textured surfaces for realism, and lighting variations (ambient and directional) to simulate diverse illumination conditions. To ensure physically plausible positioning, we utilize a physics simulation where objects are dropped into the scene and settle naturally based on their physical properties. Once the scene is stable, we define a point of interest and sample multiple camera poses around it. Each camera view follows randomized extrinsic parameters (position and orientation) while maintaining visibility of the target object. The intrinsic parameters are fixed to simulate a real-world camera. Images are then rendered from multiple angles, capturing variations in occlusion, lighting, and perspective.

\textbf{Introducing Lattice-Based Deformations}
We now modify the previously described synthetic data pipeline and introduce lattice-based deformations with the following procedure: We first place a tight lattice bounding box around the object of interest. Then we randomly perturb the control points of the lattice within an empirically set constrained range. 
Our goal is to have a unique set of annotations $\{ \mathbf{t}, \mathbf{R}, \mathbf{\delta} \}$, 
where $\mathbf{t}$ is the translation, $\mathbf{R}$ the rotation, and $\mathbf{\delta}$ the deformation.
However, the B-spline interpolation, contrary to the linear one, may introduce unwanted global translation and rotation, depending on the random pertubation.
To alleviate this amgibuity, after deforming the object, we apply the Umeyama algorithm \cite{umeyama}, and compute an optimal similarity transformation between the original mesh and the deformed one: 

    \begin{equation}
    (s, \mathbf{R}, \mathbf{t}) = \text{Umeyama}(\mathbf{P}_{\text{original}}, \mathbf{P}_{\text{deformed}}),
    \end{equation}

Here, the function \text{Umeyama} estimates the optimal similarity transformation (scaling factor \(s\) - which we fix to 1, rotation matrix \(\mathbf{R}\), and translation vector \(\mathbf{t}\)) that best aligns the set of original points \( \mathbf{P}_{\text{original}} \) with the deformed points \( \mathbf{P}_{\text{deformed}} \) in a least-squares sense. The lattice points are then corrected to ensure that deformations remain independent of pose transformations. 

\textbf{Enhancing realism with Stable Diffusion}

As mentioned in the introduction, synthetic data pipelines, like the one described so far, do not fully capture the texture variations and realism found in real-world settings. 
This limitation is particularly evident in agricultural products, where surface textures vary significantly due to ripeness, environmental conditions, and natural inconsistencies. 
Previous approaches, such as FoundationPose\cite{foundation_pose}, addressed this issue by introducing texture variations directly on 3D models before rendering. 
However, this approach still involves the rendering process, leaving a small but non-negligible domain gap between synthetic and real-world data.

In contrast, we enhance realism after rendering by applying Stable Diffusion inpainting directly to the final image, modifying only the object of interest while preserving the background and scene consistency. To ensure that the object’s pose remains unchanged, we condition the Stable Diffusion generation using ControlNet~\cite{zhang2023adding}, leveraging depth maps from the rendering pipeline. 
Finally, we curate a set of prompts describing various textures and color variations across different stages of a fruit’s lifecycle—including unripe, ripe, and rotten appearances, as well as natural color variations. This approach minimizes the domain gap while maintaining geometric and pose consistency, leading to improved generalization in real-world agricultural applications.

In summary, we aim to acquire a set of very realistic images
via the synthetic scenes, while retaining the 3D annotations required for training a network.

\subsection{Network and Training}
\label{sec:training}

Given the synthetic dataset with the 3D annotations, we can now train a neural network for detecting both the 6D pose and the deformation.
The input of the network is the cropped \textbf{RGB} (\textit{we do not use depth}) image of the desired object. During training we use the projected 2D vertices of the object to create the bounding box. For inference, we train a YOLOv11\cite{yolo11_ultralytics} model on our dataset.

For our neural network model we use a small ViT\cite{vit} backbone with $30.9M$ parameters as a feature encoder which uses 32x32 patches, pretrained on ImageNet-21k. After the ViT encoder, we employ three lightweight heads:

\noindent\textbf{Rotation Head:} predicts the 6D rotation representation \cite{r_6d_rotation} that is then orthonormalized into a valid $3\times3$ rotation matrix.

\noindent\textbf{Translation Head:} outputs a 3D vector $\mathbf{t}$ representing the object's translation in the cropped image frame.

\noindent\textbf{Deformation Head:} produces a 24D offset vector that warps a template mesh via a $2\times2\times2$ lattice.

\textbf{Training Losses.} Let $\mathbf{r}_{6d}$ be the predicted rotation (in 6D), $\mathbf{t}$ the translation, and $\delta$ the lattice offsets. We supervise them with:
\begin{itemize}
    \item \textbf{Rotation Loss:} Mean squared error (MSE) on $\mathbf{r}_{6d}$ to match the ground truth rotation.
    \item \textbf{Deformation Loss:} MSE on $\delta$, comparing them to the synthetic deformations used in data generation.
    \item \textbf{2D Projection Loss:} 
    to auxiliary supervise the estimated set of parameters $\{ \mathbf{t}, \mathbf{R}, \mathbf{\delta} \}$ and essentially derive translation parameters, we project the deformed and translated/rotated 3D vertices (i.e., the estimated final mesh) into the cropped image and measure their mean squared error against 2D keypoints of the ground-truth mesh. 
\end{itemize}

Formally,
\begin{equation}
\begin{split}
\mathcal{L} \;=\; & \lambda_{r}\,\|\mathbf{r}_{6d}^{\mathrm{pred}} - \mathbf{r}_{6d}^{\mathrm{gt}}\|^2
\;+\; \lambda_{d}\,\|\delta^{\mathrm{pred}} - \delta^{\mathrm{gt}}\|^2\\[1ex]
&\;+\; \lambda_{p}\,\|\text{Proj}(\mathbf{P} | \mathbf{R},\mathbf{t},\delta) - \mathbf{v}_{2d}^{\mathrm{gt}}\|^2.
\end{split}
\end{equation}

This combination aligns the rotation, warping, and final 2D alignment, jointly driving pose and shape accuracy. We train the network end-to-end with Adam optimizer~\cite{kingma2014adam} on cropped object patches, applying standard image augmentations to improve robustness. We train on 4,000 synthetic scenes, each with a random deformation, capturing three images per scene from different camera positions.

\subsection{From cropped to full image}
\label{subsec:crop_reconstruction}

Since the input to the network is the cropped object of interest, during inference we store the offset \(\bigl(x_{\min}, y_{\min}\bigr)\) and scale factor used for cropping and resizing. After predicting the pose in the cropped frame, we \emph{undo} these transformations to re-project the object's 2D vertices back into full-image coordinates. From there, we apply a PnP solver~\cite{lepetit2009epnp} using our lattice-deformed 3D points and the intrinsic parameters of the camera to find the translation \(\mathbf{t}\) with respect to the camera frame. 

\section{Results and Evaluation}
\textbf{Benchmark Dataset and Metrics}
To validate our framework, we collected an in-house dataset of six bananas with distinct shapes and varying ripeness stages (see top part of Fig. \ref{fig:story}). We captured 100 images using an Intel RealSense D435 and manually annotated them, after first 3D scanning each banana using an iPhone 14 Pro with a LiDAR sensor. We evaluate our method using the following metrics: (a) Chamfer Distance, which measures the geometric discrepancy between the ground truth and predicted meshes, as the scanned meshes lack per-vertex correspondence with our base banana model; (b) Mean and Median Rotation Error; (c) Mean and Median Translation Error; and (d) Deformation Error. To compute rotation and translation errors, we align the ground truth scanned meshes with the base banana mesh using ICP. The deformation error is then computed after removing the estimated rotation and translation, isolating the deformation component.

\begin{figure}
    \centering
    \includegraphics[width=1.0\linewidth]{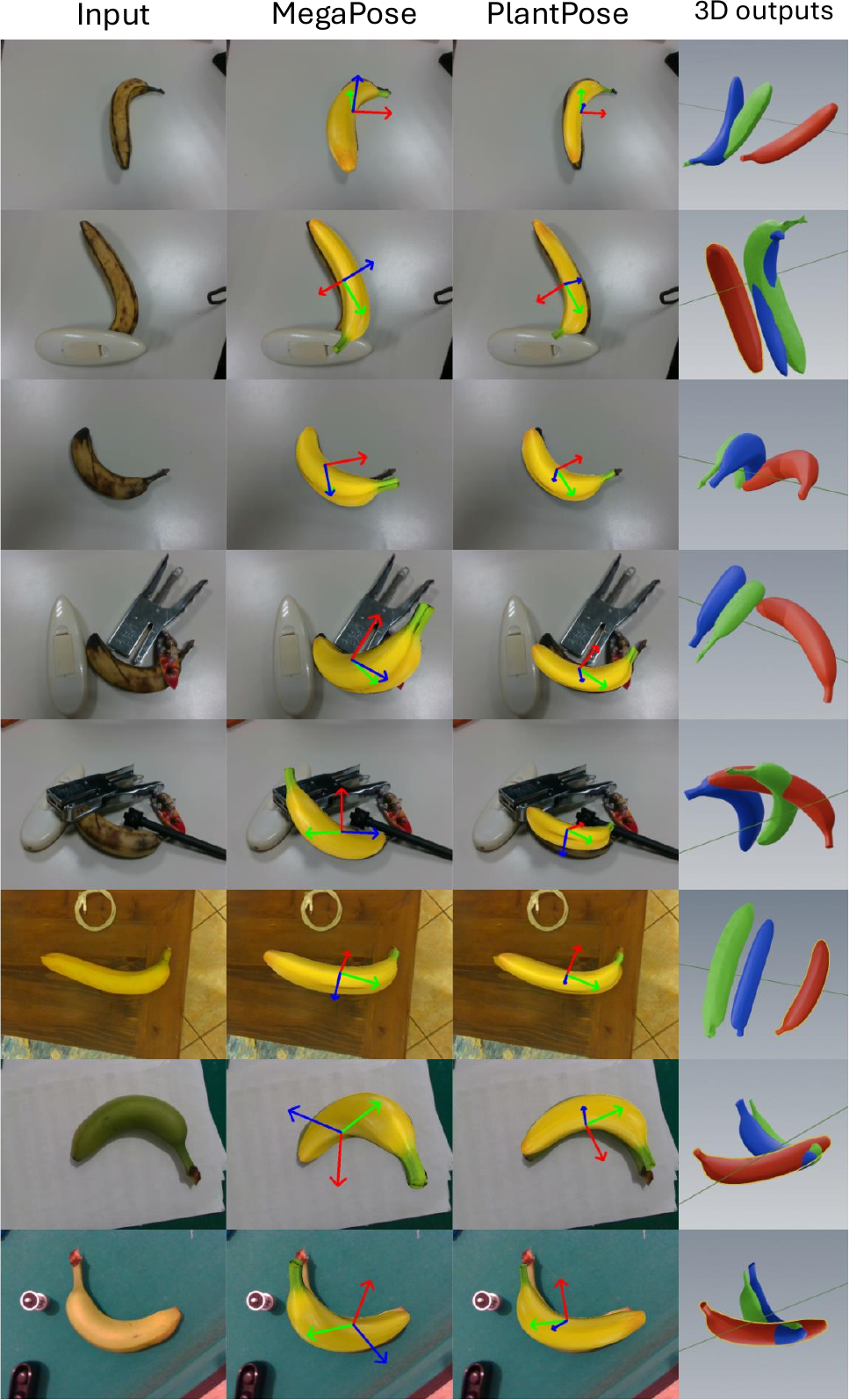}
    \caption{Qualitative results on the banana test benchmark comparing MegaPose with PlantPose. The final column shows the predictions in 3D space (with a different angle to facilitate comparison): the green color denotes the ground truth mesh and 6D pose, blue is the result of PlantPose, and Red is the result of MegaPose. }
    \label{fig:comparisons}
\end{figure}

\textbf{Comparison with State-of-the-Art.}
We compare our approach with MegaPose\cite{labbe2022megapose}, which, like our method, relies solely on RGB input. In contrast, other category-level pose estimation methods (\cite{socs,nocs,nunocs,Di_2022_CVPR,lin2021dualposenet}), to the best of our knowledge, rely on depth information (sometimes additionally to RGB). This highlights a key distinction, as our method achieves competitive performance without requiring depth, demonstrating its effectiveness as a purely vision-based solution.
We use the publicly available implementation of MegaPose and use as input the same banana mesh for all images.
We present results on our benchmark dataset in Table~\ref{tab:inhouse}. As we can see, PLANTPose significantly outperforms MegaPose across all metrics.

\begin{table}[h]
\setlength{\tabcolsep}{4pt}
    \centering
    \caption{Comparison of PLANTPose with MegaPose on the banana benchmark dataset. For all metrics, lower is better.}
    \label{tab:inhouse}
    \resizebox{\columnwidth}{!}{
    \begin{tabular}{l|c|c|c|c|c|c}
        \toprule
        Method & Chamfer (mm) $\downarrow
$ & \multicolumn{2}{c}{Rot. (deg)} $\downarrow
$ & \multicolumn{2}{c}{Trans. (mm)} $\downarrow
$ & Deform. (mm) $\downarrow
$  \\
        & Dist. & Mean & Med. & Mean & Med. & Error \\
        \midrule
        MegaPose~\cite{labbe2022megapose} & 90.1 & 52.4 & 43.6 & 59.9 & 45.6 & 29.1 \\
        PLANTPose (Ours) & \textbf{59.8} & \textbf{32.6} & \textbf{23.6} & \textbf{42.5} & \textbf{37.9} & \textbf{12.1} \\
        \bottomrule
    \end{tabular}
    }
\end{table}

Figure \ref{fig:comparisons} presents qualitative results from our test set. As shown, MegaPose struggles with accurately predicting the rotation and translation of the banana, as it relies on an average banana mesh to explain the visual scene and lacks the ability to model deformations. This limitation is particularly evident in the final column, where the results are visualized in 3D space. In contrast, PLANTPose successfully captures both the object's deformation and its 6D pose, leading to more precise translation and rotation predictions in the 3D world.

\textbf{Ablation studies}
We conduct ablation studies to evaluate the effectiveness of our Stable Diffusion-based realism enhancement and the impact of omitting the Umeyama step for correcting deformation-induced rotation and translation. The results of these experiments are presented in Table~\ref{tab:ablation}.
As evident from the results, each component plays a crucial role in achieving optimal performance.  The Umeyama correction effectively resolves ambiguities in rotation and translation introduced by the initial deformation, while the Stable Diffusion-enhanced dataset significantly improves generalization.

\begin{table}[h]
\setlength{\tabcolsep}{4pt}

    \centering
   \caption{Ablation study results evaluating the impact of different components on pose estimation accuracy. Chamfer distance (Dist.), rotation (Rot.), and translation (Trans.) errors are reported as mean (Mean) and median (Med.), along with the deformation error.}
    \label{tab:ablation}
    \resizebox{\columnwidth}{!}{
    \begin{tabular}{l|c|c|c|c|c|c}
        \toprule
        Method & Chamfer (mm) $\downarrow
$ & \multicolumn{2}{c}{Rot. (deg)} $\downarrow
$ & \multicolumn{2}{c}{Trans. (mm)} $\downarrow
$ & Deform. (mm) $\downarrow
$ \\
        & Dist. & Mean & Med. & Mean & Med. & Error \\
        \midrule
        PLANTPose (Full) & \textbf{59.8} & \textbf{32.6} & 23.6 & \textbf{42.5} & \textbf{37.9} & \textbf{12.1} \\       
        \midrule
        \quad w/o SD & 71.5 & 37.7 & \textbf{23.0} & 48.9 & 48.7 & 13.9 \\
        \quad w/o Umeyama & 73.1 & 36.5 & 26.2 & 50.3 & 45.8 & 12.7 \\
        \quad w/o Umeyama and SD & 89.7 & 36.1 & 26.7 & 61.1 & 61.9 & 26.5 \\
        \bottomrule
    \end{tabular}
    }
\end{table}

\begin{figure}
    \centering
    \includegraphics[width=1.0\linewidth]{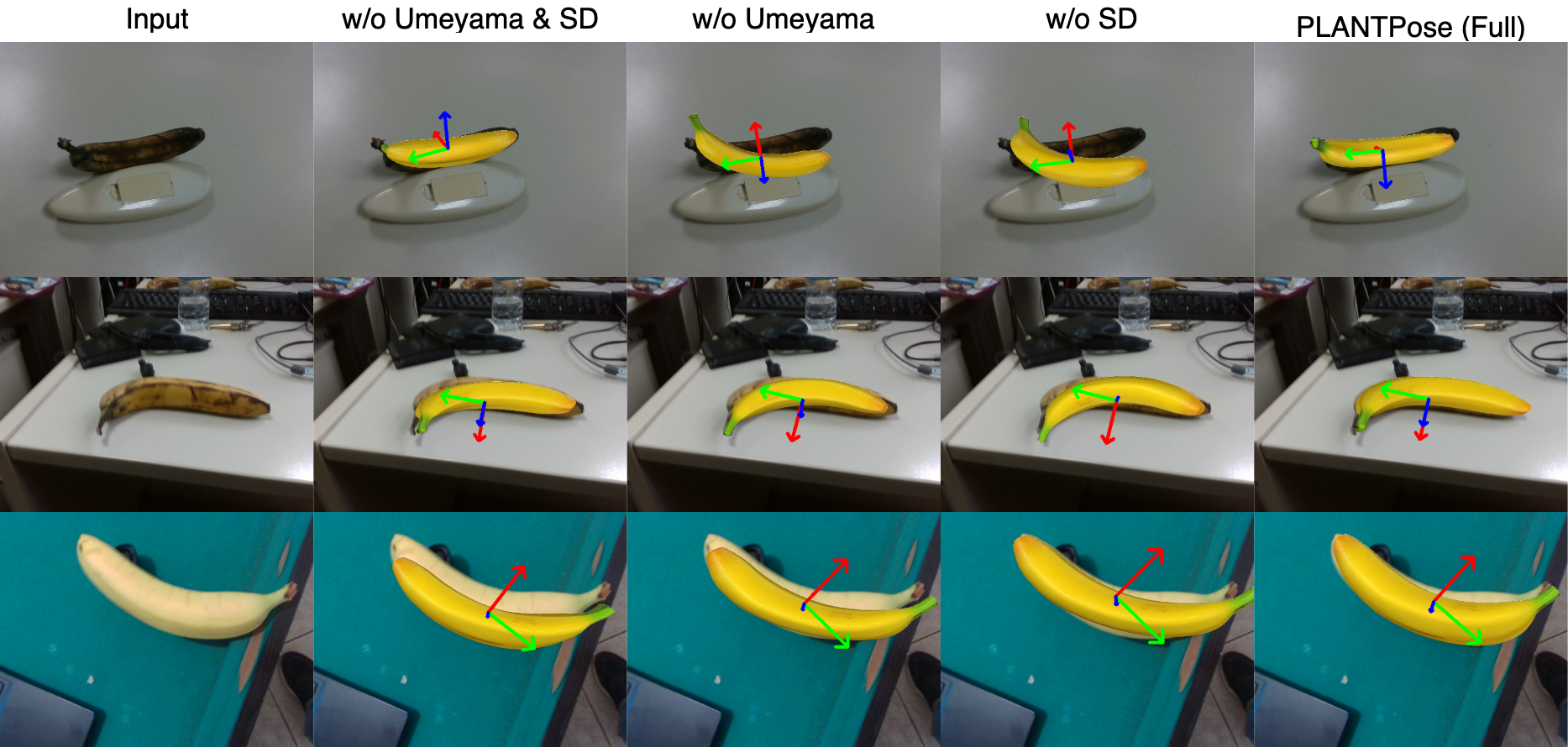}
    \caption{Qualitative ablation results from different model variations on the banana test benchmark. Each row shows a different input, while columns compare results from different models.}
    \label{fig:ablation_comparisons}
\end{figure}

Figure~\ref{fig:ablation_comparisons} shows that without Umeyama, the model struggles with translation because deformation, rotation, and translation are not disentangled—deformations inherently introduce unwanted pose changes. Without Stable Diffusion, the model performs well on familiar banana textures but fails on unseen ones, losing the overall pose.

\textbf{Limitations}
While our method accurately predicts both the deformation parameters and 6D pose of an object, it has several limitations. First, it currently requires training a separate model for each object category. A more general solution would accept a base CAD model as an additional input, adapting across multiple categories without retraining and still preserving mesh deformation capabilities. Additionally, in rare cases, even with depth conditioning, Stable Diffusion may slightly alter the object's 6D pose. Exploring more robust approaches for conditioning generative models could further enhance our synthetic data pipeline.

\section{Conclusion}

We introduced PLANTPose, a category-level 6D pose estimation framework that predicts both pose and deformation parameters from RGB images. By leveraging lattice-based deformations and Stable Diffusion-based texture augmentation, our method enables accurate pose estimation across diverse object instances without requiring instance-specific models or depth input. Our experiments demonstrate strong performance on a banana benchmark, significantly outperforming the state-of-the-art method MegaPose. For future work, we aim to expand our approach to multiple object categories beyond fruits and incorporate the base CAD model as an input, allowing greater adaptability across different object types.

\addtolength{\textheight}{-6cm}

\bibliographystyle{IEEEtran}
\bibliography{references}

\end{document}